\documentclass[11pt,a4paper]{article}
\usepackage[hyperref]{acl2019}
\usepackage{times}
\usepackage{latexsym}
\usepackage{enumitem}
\usepackage{url}
\usepackage{tablefootnote}

\aclfinalcopy 
\def\aclpaperid{073} 

\usepackage{amsmath}
\usepackage{graphicx}
\usepackage{amsfonts}
\usepackage{float}

 \usepackage{multirow}
 \usepackage[normalem]{ulem}
 \useunder{\uline}{\ul}{}
 \usepackage{booktabs} 
\usepackage{xcolor}

\author{Fangyu Liu\\
  University of Waterloo \\
  {\tt fangyu.liu@uwaterloo.ca} \\\And
  Rongtian Ye \\
  Aalto University \\
  {\tt rongtian7@gmail.com} \\ 
  }

\begin{document}


\title{A Strong and Robust Baseline for Text-Image Matching}

\maketitle

\begin{abstract}
We review the current schemes of text-image matching models and propose improvements for both training and inference. First, we empirically show limitations of two popular loss (sum and max-margin loss) widely used in training text-image embeddings and propose a trade-off: a kNN-margin loss which 1) utilizes information from hard negatives and 2) is robust to noise as all $K$-most hardest samples are taken into account, tolerating \emph{pseudo} negatives and outliers. Second, we advocate the use of Inverted Softmax (\textsc{Is}) and Cross-modal Local Scaling (\textsc{Csls}) during inference to mitigate the so-called hubness problem in high-dimensional embedding space, enhancing scores of all metrics by a large margin. 
\end{abstract}

\section{Introduction}

In recent years, deep eural models have gained a significant edge over \emph{shallow}\footnote{\emph{shallow} means non-neural methods.} models in cross-modal matching tasks. Text-image matching has been one of the most popular ones among them. Most methods involve two phases: 1) training: two neural networks (one image encoder and one text encoder) are learned end-to-end, mapping texts and images into a joint space, where vectors (either texts or images) with similar meanings are close to each other; 2) inference: for a query in modality A, after being encoded into a vector, a nearest neighbor search is performed to match the vector against all vector representations of items\footnote{In this paper, we refer to vectors used for searching as ``queries'' and vectors in the searched space as ``items''.} in modality B. As the embedding space is learned through jointly modeling vision and language, it is often referred to as \emph{Visual Semantic Embeddings} (VSE).

While the state-of-the-art architectures being consistently advanced \cite{nam2017dual,You_2018_CVPR,wehrmann2018bidirectional,Wu_2019_CVPR}, few works have focused on the more fundamental problem of text-image matching - that is, the optimization objectives during training and inference. And that is what this paper focuses on. In the following of the paper, we will discuss 1) the optimization objective during training, i.e., loss function, and 2) the objective used in inference (how should a text-image correspondence graph be predicted).

\textbf{Loss function.} \citet{faghri2018vse++} brought the most notable improvement on loss function used for training VSE. They proposed a max-margin triplet ranking loss that emphasizes on the hardest negative sample within a min-batch. The max-margin loss has gained significant popularity and is used by a big set of recent works \cite{engilberge2018finding,faghri2018vse++,lee2018stacked,Wu_2019_CVPR}. We, however, point out that the max-margin loss is very sensitive to label noise and encoder performance, and also easily overfits. Through experiments, we show that it only achieves the best performance under a careful selection of model architecture and dataset. Before \citet{faghri2018vse++}, a pairwise ranking loss has been usually adopted for text-image model training. The only difference is that, instead of only using the hardest negative sample, it sums over all negative samples (we thus refer to it as the sum-margin loss). Though sum-margin loss yields stable and consistent performance under all dataset and architecture conditions, it does not make use information from hard samples but treats all samples equally by summing the margins up. Both \citet{faghri2018vse++} and our own experiments point to a clear trend that, more and cleaner data there is, the higher quality the encoders have, the better performance the max-margin loss has; while the smaller and less clean the data is, the less powerful the encoders are, the better sum-margin loss would perform (and max-margin would fail).

In this paper, we propose the use of a trade-off: a kNN-margin loss that sums over the $k$ hardest sample within a mini-batch. It 1) makes sufficient use of hard samples and also 2) is robust across different model architectures and datasets. In experiments, the kNN-margin loss prevails in (almost) all data and model configurations.

\textbf{Inference.} During text-image matching inference, a nearest-neighbor search is usually performed to obtain a ranking for each of the queries. It has been pointed out by previous works \cite{radovanovic2010hubs,dinu2015improving,zhang2017learning} that \emph{hubs} will emerge in such high-dimensional space and nearest neighbor search can be problematic for this need. Qualitatively, the hubness problem means a small portion of queries becoming ``popular'' nearest neighbor in the search space. Hubs harm model's performance as we already know that the predicted text-image correspondence should be a \emph{bipartite matching}\footnote{In Graph Theory, a set of edges is said to be a \textbf{matching} if none of the edges share a common endpoint.}. In experiments, we show that the hubness problem is the primary source of error for inference. 
Though has not attracted enough attention in text-image matching, hubness problem has been extensively studied in Bilingual Lexicon Induction (BLI) which aims to find a matching between two sets of bilingual word vectors. We thus propose to use similar tools during the inference phase of text-image matching. Specifically, we experiment with Inverted Softmax (\textsc{Is}) \cite{smith2017offline} and Cross-modal Local Scaling (\textsc{Csls}) \cite{lample2018word} to mitigate the hubness problem in text-image embeddings.

\textbf{Contributions.} The major contributions of this work are
\begin{itemize}
    \item analyzing the shortcomings of sum and max-margin loss, proposing a kNN-margin loss as a trade-off (for training);
    \item proposing the use of Inverted Softmax and Cross-modal Local Scaling to replace naive nearest neighbor search (for inference).
\end{itemize}

\section{Method}
We first introduce the basic formulation of text-image matching model and sum/max-margin loss in \ref{sec:bf}. Then we propose our intended kNN-margin loss in Section \ref{sec:knn-loss} and the use of \textsc{Is} and \textsc{Csls} for inference in Section \ref{sec:hub_method}.

\subsection{Basic Formulation}
\label{sec:bf}
The bidirectional text-image retrieval framework consists of a text encoder and an image encoder. The text encoder is composed of word embeddings, a GRU~\cite{chung2014empirical} or LSTM~\cite{hochreiter1997long} layer and a temporal pooling layer. The image encoder is a VGG19~\cite{Simonyan14c} or ResNet152~\cite{he2016deep} pre-trained on ImageNet~\cite{deng2009imagenet} and a linear layer. We denote them as functions $f$ and $g$ which map text and image to two vectors of size $d$ respectively.

For a text-image pair $(t,i)$, the similarity of $t$ and $i$ is measured by cosine similarity of their normalized encodings:
\begin{equation}
\begin{split}
 s(i,t) = \bigg\langle  \frac{f (t)}{\| f (t) \|_2}, \frac{g (i)}{\| g (i) \|_2} \bigg\rangle : \mathbb{R}^{d} \times \mathbb{R}^{d}\rightarrow \mathbb{R}.
\end{split}
\end{equation}

During training, a margin based triplet ranking loss is adopted to cluster positive pairs and push negative pairs away from each other. We list the both the sum-margin loss used in \citet{frome2013devise,kiros2014unifying,nam2017dual,You_2018_CVPR,wehrmann2018bidirectional}:
\begin{equation}
\begin{split}
 \min_{\theta} \sum_{i\in I} \sum_{\bar{t}\in T\backslash \{t\}}  [\alpha -s(i,t)+s(i,\bar{t})]_+ \\ 
+ \sum_{t\in T} \sum_{\bar{i}\in I\backslash \{i\}}  [\alpha -s(t,i)+s(t,\bar{i})]_+;
\end{split}
\label{eq:loss_sum}
\end{equation}
and the max-margin loss used by \citet{engilberge2018finding,faghri2018vse++,lee2018stacked,Wu_2019_CVPR}:
\begin{equation}
\begin{split}
 \min_{\theta}  \sum_{i\in I} \max_{\bar{t}\in T\backslash \{t\}} [\alpha -s(i,t)+s(i,\bar{t})]_+\\ 
+ \sum_{t\in T} \max_{\bar{i}\in I\backslash \{i\}}  [ \alpha -s(t,i)+s(t,\bar{i})]_+,
\end{split}
\label{eq:loss_max}
\end{equation}
where $[\cdot]_+ =\max(0,\cdot)$; $\alpha$ is a preset margin (we use $\alpha=0.2$); $T$ and $I$ are all text and image encodings in a mini-batch; $t$ is the descriptive text for image $i$ and vice versa; $\bar{t}$ denotes non-descriptive texts for $i$ while $\bar{i}$ denotes non-descriptive images for $t$.

\subsection{kNN-margin Loss}
\label{sec:knn-loss}
We propose a simple yet robust strategy for selecting negative samples: instead of counting all (Eq.~\ref{eq:loss_sum}) or hardest (Eq.~\ref{eq:loss_max}) sample in a mini-batch, we take the $k$-hardest samples. We first define a function $\texttt{kNN}(x,M,k)$ to return the $k$ closest points in point set $M$ to $x$. Then the kNN-margin loss is formulated as:
\begin{equation}
\begin{split}
 \min_{\theta}  \sum_{i\in I} \sum_{\bar{t}\in K_1 } [\alpha -s(i,t) + s(i,\bar{t}))]_+ \\
 +\sum_{t\in T} \sum_{\bar{i} \in K_2 }  [\alpha -s(t,i) + s(t,\bar{i}))]_+
\label{eq:loss_with_prior_in_max}
\end{split}
\end{equation}
where 
$$K_1 = \texttt{kNN}(i,T\backslash\{t\},k), K_2 = \texttt{kNN}(t,I\backslash\{i\},k).$$

In max-margin loss, when the hardest sample is misleading or incorrectly labeled, the wrong gradient would be imposed on the network. We call it a \emph{pseudo} hard negative. In kNN-margin loss, though some \emph{pseudo} hard negatives might still generate false gradients, they are likely to be canceled out by the negative samples with correct information. As only the $k$ hardest negatives are considered, the selected samples are still hard enough to provide meaningful supervision to the model. In experiments, we show that kNN-margin loss indeed demonstrates such characteristics.

\subsection{Hubness Problem During Inference}
\label{sec:hub_method}
The standard procedure for inference is performing a naive nearest neighbor search. This, however, leads to the hubness problem which is the primary source or error as we will show in Section \ref{sec:hub_exp}. We thus leverage the prior that ``one query should not be the nearest neighbor for multiple items'' to improve the text-image matching. Specifically, we use two tools introduced in BLI: Inverted Softmax (\textsc{Is}) \cite{smith2017offline} and Cross-modal Local Scaling (\textsc{Csls})~\cite{lample2018word}.

\subsubsection{Inverted Softmax (\textsc{Is})}
The main idea of \textsc{Is} is to estimate the confidence of a prediction $i\rightarrow t$ not merely by similarity score $s(i,t)$, but the score reweighted by $t$'s similarity with other queries:
\begin{equation}
\begin{split}
s'(i,t) = 
\frac{e^{\beta s(i,t)}}{\sum_{\bar{i}\in I\backslash \{i\}} e^{\beta s(\bar{i},t)}}
\label{eq:IS}
\end{split}
\end{equation}
where $\beta$ is a temperature (we use $\beta=30$). Intuitively, it scales down the similarity if $t$ is also very close to other queries.

\subsubsection{Cross-modal Local Scaling (\textsc{Csls})}
\textsc{Csls} aims to decrease a query vector's similarity to item vectors lying in \emph{dense} areas while increase similarity to \emph{isolated}\footnote{\emph{Dense} and \emph{isolated} are in terms of query.} item vectors. It punishes the occurrences of an item being the nearest neighbor to multiple queries. Specifically, we update the similarity scores with the formula:
\begin{equation}
\begin{split}
s'(i,t) = 2s(i,t) - \frac{1}{k}\sum_{i_{t}\in K_1}s(i_t,t) - \frac{1}{k}\sum_{t_{i}\in K_2}s(i,t_i)
\label{eq:CSLS}
\end{split}
\end{equation}
where $K_1=\texttt{kNN}(t, I,k)$ and $K_2=\texttt{kNN}(i,T,k)$ (we use $k=10$).


\section{Experiments}
\label{sec:4.4}
\begin{table*}[h]
\small
\setlength{\tabcolsep}{4pt}
\caption{Quantitative results on Flickr30k~\cite{young2014image}. }
\label{table:f30k}
\centering
\begin{tabular}{cllcccccccccc}
\toprule
 & & \phantom{abc} & \multicolumn{5}{c}{\bf image$\rightarrow$text}  \phantom{abc} & \multicolumn{5}{c}{\bf text$\rightarrow$image} \\ \cmidrule(l){4-8} \cmidrule(l){9-13}
 \# & {\bf architecture}  & {\bf loss} &  R@1 & R@5 & R@10  & Med r    & Mean r &  \ R@1 & R@5 & R@10  & Med r & Mean r  \\   \midrule
1.1  &   \multirow{5}{*}{GRU+VGG19} &  sum-margin & 30.2 & 58.7 & 70.4 & 4.0 & 33.0 & 22.9 & 50.6 & 61.4 & \textbf{5.0} & 49.5\\ 
1.2  &  &  max-margin &  30.7 & 58.7 & 69.6 & 4.0 & 30.3 & 22.4 & 48.4 & 59.3 & 6.0 & 39.0\\
1.3 & & kNN-margin ($k=3$) & \textbf{34.1} & \textbf{61.7} & 69.9 & \textbf{3.0} & \textbf{24.7} & \textbf{25.1} & \textbf{52.5} & 64.6 & \textbf{5.0} & 34.3\\
1.4 & & kNN-margin ($k=5$) & 33.4 & 61.6 & \textbf{71.1} & \textbf{3.0} & 26.7 & 24.2 & 51.8 & \textbf{64.8} & \textbf{5.0} & \textbf{32.7} \\
1.5 & & kNN-margin ($k=10$) & 33.3 & 59.4 & 69.4 & \textbf{3.0} & 28.4 & 23.4 & 50.6 & 63.5 & \textbf{5.0} & 33.8 \\

 
\bottomrule
\end{tabular}
\end{table*}
\begin{table*}[h]
\small
\setlength{\tabcolsep}{4pt}
\caption{Quantitative results on MS-COCO~\cite{lin2014microsoft}. Using the 5k test set. }
\label{table:quant}
\centering
\begin{tabular}{cclcccccccccc}
\toprule
 & & \phantom{abc} & \multicolumn{5}{c}{\bf image$\rightarrow$text}  \phantom{abc} & \multicolumn{5}{c}{\bf text$\rightarrow$image} \\ \cmidrule(l){4-8} \cmidrule(l){9-13}
 \# & {\bf architecture}  & {\bf loss} &  R@1 & R@5 & R@10  & Med r    & Mean r &  \ R@1 & R@5 & R@10  & Med r & Mean r  \\   \midrule
 2.1  & \multirow{3}{*}{\shortstack[c]{GRU+VGG19}} &  sum-margin &  48.9 & 79.9 & 89.0 & 1.8 & 5.6 &  38.3 & 73.5 & 85.3 & \textbf{2.0} & \textbf{8.4} \\
2.2 & & max-margin & \textbf{51.8} & \textbf{81.1} & 90.5 & \textbf{1.0} & \textbf{5.5} & \textbf{39.0} & 73.9 & 84.7 & \textbf{2.0}& 12.0\\
2.3 & & kNN-margin  & 50.6 & \textbf{81.1} & \textbf{90.6} & 1.4 & \textbf{5.5} & 38.7 & \textbf{74.0} & \textbf{85.5} & \textbf{2.0} & 11.8 \\
\midrule
2.4  & \multirow{3}{*}{\shortstack[c]{ GRU+ResNet152}} & sum-margin & 53.2 & 85.0 & 93.0 & \textbf{1.0} & 3.9 & 41.9 & 77.2 & 88.0 & \textbf{2.0} & 8.7 \\
2.5  &  & max-margin & \textbf{58.7} & \textbf{88.2} & 94.0 & \textbf{1.0} & \textbf{3.2} & \textbf{45.0} & 78.9 & 88.6 & \textbf{2.0} & 8.6\\
2.6 & & kNN-margin  & 57.8 & 87.6 & \textbf{94.4} & \textbf{1.0} & 3.4 & 43.9 & \textbf{79.0} & \textbf{88.8} & \textbf{2.0} & \textbf{8.1} \\



\bottomrule
\end{tabular}
\end{table*}

In this section we introduce our experimental setups (Section~\ref{sec:4.1}, \ref{sec:4.2}, \ref{sec:4.3}) and quantitative results (Section~\ref{sec:loss_exp}, \ref{sec:hub_exp}).

\subsection{Dataset}
\label{sec:4.1}

\begin{table}[H]
\small
\caption{Train-validation-test splits of used datasets.}
\label{table:dataset}
\centering
\begin{tabular}{cccccc}
\toprule
dataset & \# train & \# validation & \# test\\
\midrule
Flickr30k & $30,000$ &  $1,000$ & $1,000$ & \\
MS-COCO 1k & $113,287$ & $5,000$ & $1,000$ \\
MS-COCO 5k & $113,287$ & $5,000$ & $5,000$ \\
\bottomrule
\end{tabular}
\end{table}
We use Flickr30k~\cite{young2014image} and MS-COCO~\cite{lin2014microsoft} as our experimental datasets. We list their splitting protocols in Table~\ref{table:dataset}. For MS-COCO, there has been several different splits used by the research community. In convenience of comparing to a wide range of results reported by other works, we use two protocols and they are referred as MS-COCO 1k and 5k where 1k and 5k differs only in the test set used (1k's test set is a subset of 5k's). Notice that MS-COCO 5k computes the average of 5 folds of 1k images.
Also, in both Flickr30k and MS-COCO, $1$ image has $5$ captions - so $5$ (text,image) pairs are used for every image.

\subsection{Evaluation Metrics}
\label{sec:4.2}
We use R@$K$s (recall at $K$), Med r and Mean r to evaluate the results: 
\begin{itemize}[noitemsep]
\item R@$K$: the ratio of ``\# of queries that the ground-truth item is ranked in top $K$'' to ``total \# of queries'' (we use $K=\{1,5,10\}$; the higher the better);
\item Med r: the median of the ground-truth ranking (the lower the better);
\item Mean r: the mean of the ground-truth ranking (the lower the better).
\end{itemize}

We compute all metrics for both text$\rightarrow$image retrieval and image$\rightarrow$text matching. We follow the convention of taking the model with maximum R@$K$s sum (both text$\rightarrow$image and image$\rightarrow$text) on the validation set as the best model for testing.

\subsection{Hyperparameters}
\label{sec:4.3}
\textbf{Training.} For max-margin models, we follow the configuration specified in \citet{faghri2018vse++}. For all other models, we start with a learning rate of 0.001 and decay it by 10 times after every 10 epochs. We train all models for 30 epochs with a batch size of $128$. All models are optimized using an Adam optimizer~\cite{kingma2014adam}.

\textbf{Model.} We use $300$-d word embeddings and $1024$ internal states for GRU text encoders (all randomly initialized with Xavier init.~\cite{glorot2010understanding}; $d=1024$ for both text and image embeddings. All image encoders are fixed (with no finetuning) for fair comparison.

\subsection{Loss Function Performance}
\label{sec:loss_exp}
Table~\ref{table:f30k} and \ref{table:quant} show quantitative results on Flickr30k and MS-COCO respectively. 

\textbf{Flickr30k.} kNN-margin loss achieves significantly better performance on all metrics than all other loss. It is worth noticing that max-margin loss fails on this dataset (even much worse than sum-margin). kNN-margin loss with $k=\{3,5\}$ get the highest scores. We use $k=3$ for the following experiments unless explicitly specified.

\textbf{MS-COCO.} Max-margin loss performs much better on MS-COCO, especially on R@$1$ - it has the best R@$1$ across both configurations. kNN-margin is comparable to max-margin. Specifically, it produces slightly worse R@$1$s, almost identical R@$5$s, and slightly better R@$10$s. Sum-margin, however, performs poorly on MS-COCO. It is worth noting that here we are using the 5k test set, which is a superset of the widely adopted 1k test set. We will compare with quantitative results reported on the 1k test set in the next section.

\subsection{Hubs during Inference}
\label{sec:hub_exp}

\begin{table*}[h]
\setlength{\tabcolsep}{2.8pt}
\caption{{Quantitative results of different inference methods across different datasets and models. Line 3.1-3.3 are using the model from Table \ref{table:f30k} line 1.3 and line 3.4-3.6, 3.15-3.17 are using the model from Table \ref{table:quant} line 2.9. Line 3.7-3.14 are results reported by previous works which all adopted naive nearest neighbor search for inference.}}
\small
\label{table:inference}
\centering
\begin{tabular}{ccclcccccccccc}
\toprule
 & & \phantom{abc} & & \multicolumn{5}{c}{\bf image$\rightarrow$text}  \phantom{abc} &  \multicolumn{5}{c}{\bf text$\rightarrow$image} \\ \cmidrule(l){5-9} \cmidrule(l){10-14}
 \#  &  {\bf dataset }  & {\bf  model } & {\bf inference} &  R@1 & R@5 & R@10  & Med r    & Mean r &  \ R@1 & R@5 & R@10  & Med r & Mean r  \\   \midrule
 3.1 & \multirow{3}{*}{Flickr30k} & \multirow{3}{*}{\shortstack[c]{GRU+VGG19 \\kNN-margin }}  & naive & 34.1 & 61.7 & 69.9 & \textbf{3.0} & 24.7 & 25.1 & 52.5 & 64.6 & 5.0 & 34.3 \\
 3.2 &  & & \textsc{Is} & \textbf{36.0} & \textbf{64.5} & \textbf{72.9} & \textbf{3.0} & \textbf{20.1} & 25.2 & 52.6 & 64.4 & 5.0 & 31.1 \\
 3.3 & & & \textsc{Csls} & \textbf{36.0} & 64.4 & 72.5 & \textbf{3.0} & 20.3 & \textbf{26.7} & \textbf{54.3} & \textbf{65.7} & \textbf{4.0} & \textbf{30.8} \\

  \midrule
  3.4 &\multirow{3}{*}{\shortstack[c]{MS-COCO \\ 5k }}  & \multirow{3}{*}{\shortstack[c]{GRU+ResNet152 \\kNN-margin }}   & naive &  57.8 & 87.6 & 94.4 & \textbf{1.0} & 3.4 & 43.9 & 79.0 & 88.8 & \textbf{2.0} & 8.1\\ 
 3.5 &  & & \textsc{Is} & \textbf{64.2} & \textbf{89.4} & 95.0 & \textbf{1.0} & 3.2 & 46.7 & 80.1 & 89.3 & \textbf{2.0} & 7.8 \\
 3.6 & & & \textsc{Csls} & 62.4 & 89.3 & \textbf{95.4} & \textbf{1.0} & \textbf{3.0} & \textbf{47.2} & \textbf{80.7} & \textbf{89.9} & \textbf{2.0} & \textbf{7.7}\\
 \midrule
  3.7  & \multirow{10}{*}{\shortstack[c]{MS-COCO \\ 1k}} & \multicolumn{2}{l}{\cite{kiros2014unifying} (ours\protect\tablefootnote{``ours'' means our implementation.})} & 49.9 & 79.4 & 90.1 & 2.0 & 5.2 & 37.3 & 74.3 & 85.9 & \textbf{2.0} & 10.8 \\ 
3.8 & &\multicolumn{2}{l}{\cite{vendrov2015order}} & 46.7 & - & 88.9  & 2.0 & 5.7 & 37.9 & - & 85.9 & \textbf{2.0} & 8.1  \\
3.9 & & \multicolumn{2}{l}{\cite{huang2017instance}}  &  53.2 & 83.1 & 91.5 & \textbf{1.0} & - & 40.7 & 75.8 & 87.4 & \textbf{2.0} & - \\
3.10 & & \multicolumn{2}{l}{\cite{liu2017learning}} & 56.4 & 85.3  & 91.5 & - & - & 43.9 & 78.1 & 88.6 & - & -\\
3.11 & & \multicolumn{2}{l}{ \cite{You_2018_CVPR} } &  56.3 & 84.4 & 92.2 & \textbf{1.0} & - & 45.7 & 81.2 & 90.6 & \textbf{2.0} & - \\
3.12 & & \multicolumn{2}{l}{\cite{faghri2018vse++}  } & 58.3 & 86.1 & 93.3 &  \textbf{1.0} & - & 43.6 & 77.6 & 87.8 & \textbf{2.0} \\
3.13 & & \multicolumn{2}{l}{\cite{faghri2018vse++} (ours) }& 60.5 & 89.6 & 94.9 & \textbf{1.0} & 3.1 & 46.1 & 79.5 & 88.7 & \textbf{2.0} & 8.5 \\
 3.14 & & \multicolumn{2}{l}{\cite{Wu_2019_CVPR}} & 64.3 & 89.2 & 94.8 & \textbf{1.0} & - & 48.3 & 81.7 & \textbf{91.2} & \textbf{2.0} &  - \\
 3.15 &   & \multirow{3}{*}{\shortstack[c]{GRU+ResNet152 \\kNN-margin }} & naive &  58.3 & 89.2 & 95.4 & \textbf{1.0} & 3.1 & 45.0 & 80.4 & 89.6 & \textbf{2.0} & 7.2\\
 3.16 & & &\textsc{Is} & \textbf{66.4} & 91.8 & 96.1 & \textbf{1.0} & 2.7 & 48.6 & 81.5 & 90.3 & \textbf{2.0} & 7.3\\
 3.17 & & &\textsc{Csls} &  65.4 & \textbf{91.9} & \textbf{97.1} & \textbf{1.0} & \textbf{2.5} & \textbf{49.6} & \textbf{82.7} & \textbf{91.2} & \textbf{2.0} & \textbf{6.5} \\
\bottomrule
\end{tabular}
\end{table*}

To show hubness is indeed a major source of error, we select one of the text-image embeddings to do statistics. We use the model on Table~\ref{table:quant} line 2.1 to generate embeddings on MS-COCO's test set. Among the $25,000$ (query, item) pairs, only $1,027$ ($4.1\%$) items are the nearest neighbor (NN) of solely $1$ query; there are, however, $19,805$ ($79.2\%$) items that are NN to $0$ query and $3,007$ ($12.0\%$) items that are NN to $\ge 5$ queries, indicating wide existence of hubs. Moreover, the most ``popular'' item is NN to $51$ queries. We know that one item ought to be NN to only one query in the ground-truth query-item matching. So, we can spot errors even before ground-truth labels are revealed - for instance, the most ``popular'' item with $51$ NNs must be the \emph{false} NN for at least $50$ queries. Table~\ref{table:hub} shows the brief statistics.

\begin{table}[H]
\small
\setlength{\tabcolsep}{4pt}
\caption{Statistics of \# items being NN to $k$ queries in the embeddings of Table~\ref{table:quant}, line 2.1, text$\rightarrow$image. There are in total 25,000 (text,image) paris in this embedding.}
\label{table:hub}
\centering
\begin{tabular}{cccccc}
\toprule
& $k=0$ & $k=1$ & $k\ge2$ & $k\ge5$  & $k\ge 10$\\
\midrule
\# & 19,805  & 1,026  & 4,169  & 3,007  & 500 \\
percentage & 79.2\% & 4.1\% & 16.7\% & 12.0\% & 2.0\%\\
\bottomrule
\end{tabular}
\end{table}

 Both \textsc{Is} and \textsc{Csls} demonstrate compelling empirical performance in mitigating the hubness problem. Table~\ref{table:inference} shows the quantitative results. R@$K$s and also Med r, Mean r are improved by a large margin with both methods. In most configurations, \textsc{Csls} is slightly better than \textsc{Is} on improving text$\rightarrow$image inference while \textsc{Is} is better at image$\rightarrow$text. The best results (line 3.8, 3.9) are even better than the recently reported state-of-the-art \cite{Wu_2019_CVPR} (Table \ref{table:inference} line 3.14), which performs a naive nearest neighbor search. This suggests that the hubness problem deserves much more attention and careful selection of inference methods is vital for text-image matching.

\section{Limitations and Future Work}
This paper brings up a baseline with excellent empirical performance. We plan to contribute more theoretical and technical novelty in follow up works for both the training and inference phase of text-image matching models.

\textbf{Loss function.} Though the kNN-margin loss has superior empirical performance, it is leveraging the prior knowledge we hardcoded in it - it relies on a suitable $k$ to maximize its power. Flickr30k and MS-COCO are relatively clean and high-quality datasets while the real world data is usually not. With the kNN-margin loss being a strong baseline, we plan to bring a certain form of self-adaptiveness into the loss function to help it automatically decide what to learn based on the distribution of data points.

Also, to further validate the robustness of loss functions, we plan to experiment models on more \emph{noisy} data. The reason for max-margin's failure on Flikr30k is more likely that the training set is too small - so the model easily overfits. However, the dataset (Flikr30k) itself is rather clean and accurate. It makes more sense to experiment with a noisy dataset with \emph{weak} text-image correspondence or even false labels. We have two types of candidates for this need: 1) academic datasets that contain ``foil'' \cite{shekhar2017foil_acl} or adversarial samples \cite{shi2018learning}; 2) a real-world text-image dataset such as a news article-image dataset~\cite{elliott20161,biten2019good}.

\textbf{Inference.} Both \textsc{Is} and \textsc{Csls} are \emph{soft} criteria. If we do have the strong prior that the final text-image correspondence is a bipartite matching, we might as well make use of that information and impose a \emph{hard} constraint on it. The task of text-image matching, after all, is also a form of assignment problem in Combinatorial Optimization (CO). We thus plan to investigate tools from the CO literature such as the Hungarian Algorithm~\cite{kuhn1955hungarian}, which is the best-known algorithm for producing a maximum weight bipartite matching; the Murty's Algorithm \cite{murty1968letter}, which generalizes the Hungarian Algorithm into producing the $K$-best matching - so that rankings are available for computing R@$K$ scores.

\section{Related Work}
In this section, we introduce works from two fields which are highly-related to our work: 1) text-image matching and VSE; 2) Bilingual Lexicon Induction (BLI) in the context of cross-modal matching.

\subsection{Text-image Matching}
 Since the dawn of deep learning, works have emerged using a two-branch structure to connect both language and vision. \citet{frome2013devise} brought up the idea of VSE, which is to embed pairs of (text, image) data and compare them in a joint space. Later works extended VSE for the task of text-image matching \cite{hodosh2013framing,kiros2014unifying,gong2014improving,vendrov2015order,tsai2017learning,faghri2018vse++,wang2019learning}, which is also our task of interest. It is worth noting that there are other lines of works which also jointly model language and vision. The closest one might be image captioning \cite{lebret2015phrase,karpathy2015deep}. But image captioning aims to generate novel captions while text-image matching retrieves existing descriptive texts or images in a database.
 
 
 
\subsection{Bilingual Lexicon Induction (BLI)}
\label{sec:bli}
We specifically talk about BLI as the tools we used to improve inference performance come from this literature. 
BLI is the task of inducing word translations from monolingual corpora in two languages~\cite{irvine2017comprehensive}. Words are usually represented by vectors trained from Distributional Semantics, eg. \citet{mikolov2013distributed}. So, the word translation problem converts to finding the appropriate matching among two sets of vectors which makes it similar to our task of interest.
\citet{smith2017offline,lample2018word} proposed to first conduct a direct Procrustes Analysis \cite{schonemann1966generalized} between two sets of vectors, then use criteria that heavily punish hubs during inference to avoid the hubness problem. We experimented with both methods in our task.

\section{Conclusion}

We discuss the pros and cons of prevalent loss functions used in text-image matching and propose a kNN-margin loss as a trade-off which yields strong and robust performance across different model architectures and datasets. Instead of using naive nearest neighbor search, we advocate to adopt more polished inference strategies such as Inverted Softmax (\textsc{Is}) and Cross-modal Local Scaling (\textsc{Csls}), which can significantly improve scores of all metrics. 

We also analyze the limitations of this work and indicate the next step for improving both the loss function and the inference method.

\section{Acknowledgement}
We thank the reviewers for their careful and insightful comments. We thank our family members who have both spiritually and financially supported our independent research. The author Fangyu Liu thanks R\'emi Lebret for introducing him this interesting problem and many of the related works.

\bibliography{acl2019}
\bibliographystyle{acl_natbib}

\appendix



\end{document}